\documentclass[pdflatex,sn-basic]{sn-jnl}
\usepackage{amsmath,amsfonts,bm}
\usepackage{algorithmic}
\usepackage{algorithm}
\usepackage{array}
\usepackage{subcaption}
\usepackage{textcomp}
\usepackage{url}
\usepackage{csquotes}
\usepackage{verbatim}
\usepackage{graphicx}
\usepackage[dvipsnames]{xcolor}

\DeclareMathOperator*{\argmin}{arg\!min}

\begin{document}

\def\tr{\top}
\def\ov{\overline}
\def\til{\widetilde}
\def\lla{\left\langle}
\def\rra{\right\rangle}

\def\x{{\mathbf x}}
\def\c{{\mathbf c}}
\def\p{{\mathbf p}}
\def\w{{\mathbf w}}
\def\m{{\mathbf m}}
\def\a{{\mathbf a}}
\def\b{{\mathbf b}}
\def\g{{\mathbf g}}
\def\y{{\mathbf y}}
\def\z{{\mathbf z}}
\def\f{{\mathbf f}}
\def\s{{\mathbf s}}
\def\d{{\mathbf d}}
\def\n{{\mathbf n}}
\def\h{{\mathbf h}}
\def\z{{\mathbf z}}
\def\t{{\mathbf t}}
\def\r{{\mathbf r}}
\def\w{{\mathbf w}}
\def\u{{\mathbf u}}
\def\q{{\mathbf q}}
\def\v{{\mathbf v}}
\def\e{{\mathbf e}}
\def\q{{\mathbf q}}

\def\ti{{\mathbf {\tilde t}}}
\def\qi{{\mathbf {\tilde q}}}

\def\D{{\mathbf D}}
\def\M{{\mathbf M}}
\def\P{{\mathbf P}}
\def\R{{\mathbf R}}
\def\H{{\mathbf H}}
\def\l{{\mathbf L}}
\def\B{{\mathbf B}}
\def\E{{\mathbf E}}
\def\Z{{\mathbf Z}}
\def\A{{\mathbf A}}
\def\X{{\mathbf X}}
\def\G{{\mathbf G}}
\def\C{{\mathbf C}}
\def\Y{{\mathbf Y}}
\def\T{{\mathbf T}}
\def\W{{\mathbf W}}
\def\Q{{\mathbf Q}}
\def\K{{\mathbf K}}
\def\U{{\mathbf U}}
\def\V{{\mathbf V}}
\def\I{{\mathbf I}}
\def\R{{\mathbf R}}
\def\S{{\mathbf S}}

\def\cR{{\cal R}}
\def\cT{{\cal T}}	
\def\cM{{\cal M}}
\def\cN{{\cal N}}
\def\cL{{\cal L}}
\def\cW{{\cal W}}
\def\cX{{\cal X}}
\def\cY{{\cal Y}}
\def\cZ{{\cal Z}}
\def\cD{{\cal D}}
\def\cB{{\cal B}}
\def\cS{{\cal S}}
\def\cV{{\cal V}}
\def\cG{{\cal G}}
\def\cE{{\cal E}}
\def\cC{{\cal C}}
\def\cQ{{\cal Q}}
\def\cK{{\cal K}}
\def\cA{{\cal A}}
\def\cO{{\cal O}}
\def\cU{{\cal U}}
\def\cI{{\cal I}}
\def\cF{{\cal F}}

\newcommand{\RR}{{\mathbb{R}}}
\newcommand{\CC}{{\mathbb{C}}}
\newcommand{\EE}{{\mathbb{E}}}
\newcommand{\NN}{{\mathbb{N}}}
\newcommand{\ZZ}{{\mathbb{Z}}}

\newcommand*\diff{\mathop{}\!\mathrm{d}}

\newcommand{\bigslant}[2]{{\raisebox{.2em}{$#1$}\left/\raisebox{-.2em}{$#2$}\right.}}

\newcommand{\cii}{[C~\textsc{ii}]}

\title{Discriminative Subspace Emersion from learning feature relevances across different populations}

\author*[1]{\fnm{Marco} \sur{Canducci}}\email{M.Canducci@bham.ac.uk}\equalcont{These authors contributed equally to this work.}

\author[2]{\fnm{Lida} \sur{Abdi}}\email{l.abdi@lms.mrc.ac.uk}\equalcont{These authors contributed equally to this work.}

\author[3,4]{\fnm{Alessandro} \sur{Prete}}\email{a.prete@bham.ac.uk}

\author[2,5]{\fnm{Roland J.} \sur{Veen}}\email{r.j.veen@lms.mrc.ac.uk,r.j.veen@rug.nl}

\author[5,6]{\fnm{Michael} \sur{Biehl}}\email{m.biehl@rug.nl}

\author[2,7]{\fnm{Wiebke} \sur{Arlt}}\email{w.arlt@lms.mrc.ac.uk}

\author[1]{\fnm{Peter} \sur{Tino}}\email{p.tino@bham.ac.uk}

\affil*[1]{\orgdiv{School of Computer Science}, \orgname{University of Birmingham}, \orgaddress{ \city{Birmingham}, \postcode{B15 2TT}, \country{UK}}}

\affil[2]{\orgdiv{Medical Research Council Laboratory of Medical Sciences, Institute of Clinical Sciences}, \orgname{Imperial College London}, \orgaddress{\city{London}, \postcode{W12 0HS}, \country{UK}}}

\affil[3]{\orgdiv{Department of Metabolism and Systems Science, School of Medical Sciences, College of Health and Medicine}, \orgname{University of Birmingham, Birmingham}, \orgaddress{\city{Birmingham}, \postcode{B15 2TT}, \country{UK}}}

\affil[4]{\orgdiv{NIHR Birmingham Biomedical Research Centre}, \orgname{University of Birmingham and University Hospitals Birmingham NHS Foundation Trust}, \orgaddress{\city{Birmingham}, \postcode{B15 2TT}, \country{UK}}}

\affil[5]{\orgdiv{Bernoulli Institute for Mathematics and Artificial Intelligence}, \orgname{ University of Groningen}, \orgaddress{\city{Groningen}, \postcode{1022 9701}, \country{NL}}}

\affil[6]{\orgdiv{Centre for Systems Modelling and Quantitative Biomedicine (SMQB)}, \orgname{University of Birmingham}, \orgaddress{\city{Birmingham}, \postcode{B15 2TT}, \country{UK}}}

\affil[7]{\orgdiv{Institute of Clinical Sciences}, \orgname{Imperial College London}, \orgaddress{\city{London}, \postcode{W12 0HS}, \country{UK}}}

\maketitle

\begin{abstract}

In a given classification task, the accuracy of the learner is often hampered by finiteness of the training set, high-dimensionality of the feature space and severe overlap between classes. In the context of interpretable learners, with (piecewise) linear separation boundaries, these issues can be mitigated by careful construction of optimization procedures and/or estimation of relevant features for the task. However, when the task is shared across two disjoint populations the main interest is shifted towards estimating a set of features that discriminate the most between the two, when performing classification.
We propose a new Discriminative Subspace Emersion (DSE) method to extend subspace learning toward a general relevance learning framework. DSE allows us to identify the most relevant features in distinguishing the classification task across two populations, even in cases of high overlap between classes. The proposed methodology is designed to work with multiple sets of labels and is derived in principle without being tied to a specific choice of base learner. 
Theoretical and empirical investigations over synthetic and real-world datasets indicate that DSE accurately identifies a common subspace for the classification across different populations. This is shown to be true for a surprisingly high degree of overlap between classes.  
\end{abstract}


\section{Introduction}
\label{sec:Introduction}

In supervised learning, annotated data is used to extract information about the properties that most distinguish the classes in which data is organized. In this sense, classification generally aims at recovering a boundary that separates the given classes. It is often good practice to assume that not all features are equally relevant in this discrimination, but some display large variability across the different data categories. The goal of finding a subset of relevant features for the classification task is often referred to as subspace learning. 

Extreme examples of this approach are Support Vector Machine (SVM, \cite{cortes1995support}) and logistic regression. SVM performs linear separation by finding the optimal hyperplane that separates the classes in the input space. Logistic regression performs linear separation by modelling the probability that a given input belongs to a particular class. A different category of supervised classification algorithms approach the discrimination between the classes via the use of prototypes, encapsulating and compressing the properties of the classes into a small set of typical examples. Examples of prototype-based methodologies are the different implementations of Learning Vector Quantization (LVQ, \cite{Kohonen1995SelfOrganizingM}). 

Instead of finding a subset of meaningful features for the classification, metric learning methodologies (\cite{MAL-019} and references therein) focus on determining a metric tensor that endows the feature space with a geodesic distance. By pushing similar points (members of the same class) closer and dissimilar points (members of different classes) further apart, this distance acts as a dissimilarity measure. When the constraints of the dissimilarity matrix are relaxed, in particular positive definiteness, the learning process might discard irrelevant features in the classification task, de facto extracting a meaningful subset that defines a subspace of the original space. Although in this case the dissimilarity matrix is not exactly a metric tensor, its reduction to the lower-dimensional subspace is. Thus, slightly abusing notation, in the following we will refer to this matrix as the metric tensor.

Generalized Matrix Learning Vector Quantization (GMLVQ, \cite{schneider2009adaptive}) is a prototype-based classification algorithm that also estimates a dissimilarity metric from the data. GMLVQ aims to achieve piecewise linear separation between classes by adjusting the prototype vectors and the relevance matrix such that data points from different classes are well-separated in the feature space. GMLVQ has successful applications in various domains such as pattern recognition, image classification, and bio-informatics \cite{biehl2013analysis,van2020application}. The adaptive metric tensor accounts for the correlation between the features, and it provides information about the structure of the data. The diagonal elements of the metric tensor indicate the importance of features (\emph{relevances}) and off-diagonal elements indicate correlation and dependencies of the features. It implicitly identifies important features through the learned discriminative subspace. 

Discriminative subspace learning focuses on projecting high-dimensional data into a lower-dimensional space while maintaining class separability. \cite{chen_supervised_2025} propose a supervised dimension reduction method using linear projection and Kullback-Leibler divergence to optimize feature separability. \cite{vogelstein_supervised_2021} extend this approach to big data, leveraging scalable algorithms for large datasets. \cite{yin_discriminative_2023} employ Riemannian manifold optimization for enhanced class discrimination. 

\cite{fu_subspace_2022, fu_classification_2023} develop and evaluate the Subspace Learning Machine (SLM), incorporating decision trees and nonhomogeneous media analysis for optimized classification. 
\cite{amiri_subspace_2025} introduce a subspace aggregating algorithm combining bagging, boosting, and random forests for enhanced classification accuracy. \cite{yan_novel_2006} present a scalable supervised subspace learning algorithm, while \cite{yan_graph_2007} establish a graph embedding framework that generalizes various dimensionality reduction methods.  

\cite{dwivedi_linear_2021} analyze linear discriminant analysis (LDA, \cite{hastie01statisticallearning}) under f-divergence measures to enhance stability. \cite{fukui_discriminant_2023} introduce generalized difference subspaces for discriminative feature extraction, expanding on prior work by  \cite{fukui_difference_2015} on difference subspaces and their applications in subspace-based methods. \cite{ren_commonality_2024} propose a commonality and individuality-based subspace learning approach, integrating multitask learning principles. 
However, in some applications, identification of important and relevant features is harmed by high overlap between classes. To complicate this scenario, many classification (or subspace learning) algorithms include a level of stochasticity, given by either random initialization or the chosen optimization procedure. Not to mention that if classes are imbalanced, subsampling needs to be adopted to obtain unbiased classifiers. In order to mitigate the effects of these sources of uncertainty, ensemble approaches to classification \cite{Sollich_Krogh_1995,Zhou_2012}, also referred to as mixture of experts \cite{Masoudnia_MoE_2014}, can be used \cite{Zahavy2016EnsembleRA}. 

Evidently, a plethora of Subspace Learning algorithms exists that rely on geometrical or informational theoretical notions for identifying a discriminative subspace for the classification task at hand. But when the same classification task needs to be performed over two distinct populations, the focus must be shifted from learning the individual optimal discriminative subspaces, towards identifying the common subspace where populations differ the most (under the lens of the specific classification task). 
Note that this is different from addressing a multi-label classification problem (e.g. \cite{tarekegn_MLC_2021}) and the corresponding discriminative subspace estimation \citep{bayati_mssl_2022,ma_discriminative_2024}.

To better highlight the scenario proposed in this work, consider the case where the same classification task (presence or not of a health condition in a cohort) has to be performed over two different populations (e.g. patients with different degrees of Cortisol excess). In the following, different populations will be identified by letters ``A'' and ``B'' and different health states by ``Condition'' and ``No condition''. We refer to Phase 1 - Case 1 when performing the classification task over the Conditions in Population A and Phase 1 - Case 2 when the same classification is performed in Population B.
When the overlap between classes in both populations is high, the classification performances are bound to be poor and feature relevances noisy in both cases. Thus, training a number of classifiers in each Case yields a distribution over the relevance of features. Although the classification performance in both cases might approach random guessing, the (noisy) relevances might have captured information useful in the discrimination of the same classification task across the two populations. Having now samples of the relevances for each Case, it is possible to identify the relevant features that distinguish between the populations by again applying a classifier to the two sets of relevances. Performing the classification of feature relevances is what we refer to as Phase 2.

Even when the classification performance is extremely poor in Phase 1 (high overlap between classes) we verify that the proposed methodology can identify the separation direction with high confidence, uncovering information deeply buried in the two Cases of Phase 1. To the best of our knowledge, this is the first study that approaches such a problem. Furthermore, our proposed methodology is independent from the choice of subspace/feature relevance learning methodology but we choose to demonstrate it by comparing the results of GMLVQ and SVM as base learners, due to their interpretability.

The remainder of the paper is organized as follows: Section \ref{sec:Subspace_Learning_Algorithms_in_Classification_Context} provides background information about the subspace learning methods used in the proposed methodology. It also provides details regarding the feature relevance in subspace learning. Section \ref{sec:First_Intuitions_of_Double_Subspace_Learning} presents intuitions about the proposed method and provides some illustrative examples. In Section \ref{sec:Double_Subspace_Learning_Theoretical_Motivations}, the proposed methodology is explained in details. Experimental settings, results, and analysis of the results are provided in Section \ref{sec:Experimental_Settings}. Section \ref{Conclusion} concludes the paper.

\section{Subspace Learning Algorithms in Classification Context}
\label{sec:Subspace_Learning_Algorithms_in_Classification_Context}
There are many examples of subspace learning algorithms in the literature; however, in this study we focus on GMLVQ and SVM. Following subsections provide essential background information about these methods and teir connection to subspace learning.

\subsection{Generalized Matrix Learning Vector Quantization (GMLVQ)}
GMLVQ generalizes Learning Vector Quantization (LVQ) \citep{Kohonen1995SelfOrganizingM} by adopting a different notion of distance. While LVQ uses Euclidean distance, GMLVQ applies a similarity measure in terms of a symmetric and positive-definite matrix $\bm{\Lambda} = \bm{\Omega}^\top \bm{\Omega}$, through which an inner product is defined. $\bm{\Omega}$ is a $d \times d$ arbitrary matrix. The distance takes the form:
\begin{equation}
	d^{\bm{\Lambda}} (\w, \x) = ( \x - \w)^T\bm{\Lambda} (\x - \w)                        
\end{equation}
Positive definiteness can be imposed by enforcing $\mathrm{det}(\bm{\Lambda}) \neq 0$. 
The optimization of the algorithm is obtained by minimization of the cost function:
\begin{equation}
f = \sum_{i=1}^N \Phi(\mu(\x_i)), \quad \mu(\x_i) = \frac{d^{\Lambda} (\w^p,\x_i) - d^{\Lambda} (\w^q,\x_i) }{d^{\Lambda} (\w^p,\x_i) + d^{\Lambda} (\w^q,\x_i)}
\end{equation}
where $\Phi$ is a monotonic function (e.g. the logistic function) and $\x_i,  \forall \,i=1,\dots,n$ are samples from the dataset.
As derived in \cite{Hammer2005SupNeuralGas} and later adapted in \cite{schneider2009adaptive}, the update equations for the prototypes and the elements of matrix $\bm{\Omega}$, at a given iteration $t$ and for a single sample $\x$, can be analytically derived. 

Qualitatively, the update equations imply that the closest correct prototype is pulled towards sample $\x$ while the closest incorrect is pushed away from it. At the same time, the distance from the closest correct prototype is decreased, while it is increased for the closest incorrect prototype.
In order to avoid degeneration, after each update in GMLVQ, a normalization over $\bm{\Lambda}$ is imposed so that $tr{(\bm{\Lambda}}) = 1$.

The sum of diagonal elements coincides with the sum of eigenvalues which generalizes the normalization of relevances $\sum_i \bm{\Lambda}_i = 1$ for a simple diagonal metric.
GMLVQ can be used for subspace learning, and it is particularly suitable for problems in which the data lie on or near a lower-dimensional subspace. By learning a set of prototypes and metric tensor $\bm{\Lambda}$, GMLVQ combines both aspects of vector quantization and subspace learning to classify data points and project them into a lower-dimensional subspace (which maximizes the separation). The metric tensor essentially defines how each feature contributes to the classification task. Projecting the samples to a lower dimensional space reduces the noise and redundancy in the data, and it helps to focus on the information that is most useful for classification.  

\subsection{Support Vector Machine (SVM)}
In two-class problems, the goal of a classifier is to find the decision boundary that separates the two classes. If the classes are separable in some feature space (given by a chosen kernel), the decision boundary is a linear hyperplane. If such a hyperplane exists, it is identified by only a subset of the training set; the support vectors. From this notion, comes the methodology of Support Vector Machines (SVMs). These are classifiers whose aim is to find the optimal hyperplane to separate the two classes \cite{Vapnik2006EstimationOD}. The decision over the label of a new observation $\x$ is given by the sign of:
\begin{equation}
y(\x) = \bm{\omega}^\top K(\x) + \b
\end{equation}
where $\bm{\omega}$ is the perpendicular vector to the hyperplane and $\b$ the bias. In SVM, it is generally assumed that $c_1 = -1$ and $c_2 = +1$ for a training point, or $\ell(\x_i) \in \{-1,+1\}$ for $i=1,\dots,n$. Given this assumption, when classes are linearly separable in feature space, we have $y(\x_i)>0$ for $\ell(\x_i) = +1$ and $y(\x_i)<0$ for $\ell(\x_i) = -1$, so that $\ell(\x_i)y(x_i) > 0 \, \forall \, i=1,\dots,n$. Here, the distance of the closest support vector to the hyperplane, along its perpendicular direction, is the \emph{margin}. For separable classes, SVMs are said to find a \emph{hard margin} by solving the optimization problem:
\begin{equation}
\argmin_{\bm{\omega},\b} \frac{1}{2}\|\bm{\omega}\|^2
\end{equation}
under the constraint $\ell(\x_i)(\bm{\omega}^\top K(\x_i) + \b) \geq 1, \forall \, i=1,\dots,n$.

However, when classes are not separable the hard margin formulation of SVM cannot be used, because no hyperplane exists that can separate the classes without committing errors. To solve this problem a \emph{soft margin} formulation has been proposed \cite{cortes1995support,Bennet_SVM_1992}. The optimization problem can be rewritten by relaxing the missclassification penalty for a given sample, via the use of \emph{slack} variables. 
Again, the orthogonal vector to the hyperplane, $\bm{\omega}$, points to the separation direction of the two considered classes. Hence, SVM can be considered as a robust one dimensional subspace learning algorithm in binary classification problems.

\subsection{From Learning Subspace to Feature Relevance}
\label{subsec:From_Learning_Subspace_to_Feature_Relevance}
Through the learning process, GMLVQ learns both the relevance matrix $\bm{\Lambda}$ and the prototypes. Eigendecomposition of $\bm{\Lambda}$ provides its eigenvalues $\tilde\sigma_k$ and eigenvectors $\bm{v}_{k}$, for $k=1,\dots,d$. Having defined $\bm{\Lambda}$ as a positive-semi-definite matrix, the number of non-negative eigenvalues is always $d$.
Through its eigendecomposition, matrix $\bm{\Lambda}$ can be written as:
\begin{equation}\label{eq:LambdaEigendec}
\bm{\Lambda} = \sum_{k=1}^d \tilde\sigma_k (\v_{k} \v_{k}^\top)
\end{equation}
By definition, the relevance $r_j$ of feature $j$ is the $j-$th diagonal element of matrix $\bm{\Lambda}$, and via equation \eqref{eq:LambdaEigendec} it can be written as:
\begin{equation}
	r_j = \sum_{k=1}^d \tilde\sigma_k (\v_{j,k})^2
\end{equation}
where index $k$ and $j$ identify the eigenvector and the co-ordinate of the considered feature, respectively.

The eigenvectors of $\bm{\Lambda}$ form a new orthonormal basis for $\RR^d$. Despite matrix $\bm{\Lambda}$ being generally full-rank, the eigenvalues $\tilde\sigma_k$ indicate the importance of each eigenvector in determining the distance in the subspace. Only the dominant eigenvectors play a significant role in the estimation. Through this basis and the corresponding eigenvalues, each feature is assigned a weight (relevance) representing its importance in the classification problem. Features with a higher weight are considered more relevant than features with a lower weight. In this sense, GMLVQ discovers the dominant directions spanning the subspace where the classification has higher performance. The directions are combinations of input features (rotated axis) and can be interpreted as the dominant degrees of freedom driving the classification task. Furthermore, the relevances can be used for feature selection since irrelevant features can be discarded in further analysis or label estimation for unseen data to reduce the complexity. This represents a subspace learning procedure in the original feature axis system.

While GMLVQ recovers a high rank metric tensor, describing the subspace that maximizes the separation between classes, SVM compresses the classification task into the direction $\bm{\omega}$. As stated above, $\bm{\omega}$ is the perpendicular vector to the separating plane. Assuming that the only meaningful direction for separation between classes is equivalent to stating that the metric tensor is rank one, with only one non-zero eigenvector $\bm{\omega}$. In this sense, a full basis for the subspace can still be identified, but the eigenvalues for all vectors are zeros except the one for $\bm{\omega}$ which is $1$. Equation \eqref{eq:LambdaEigendec} can, then, be rewritten as:
\begin{equation}\label{eq:LambdaSVM}
\bm{\Lambda} = \sum_{k=1}^d \tilde\sigma_k (\v_{k} \v_{k}^\top) = 1\,\bm{\omega} \bm{\omega}^\top + 0 + \dots + 0 = \bm{\omega}\bm{\omega}^T
\end{equation}
This implies that the relevances for the input features are the squared coordinates of vector $\bm{\omega}$.

\section{First Intuitions}
\label{sec:First_Intuitions_of_Double_Subspace_Learning}
\begin{figure*}[ht!]
	\centering
	\subfloat[][]{\label{subfig:Intuition}\includegraphics[width = \textwidth,trim=0.7cm 0.6cm 1.8cm 0.2cm,clip]{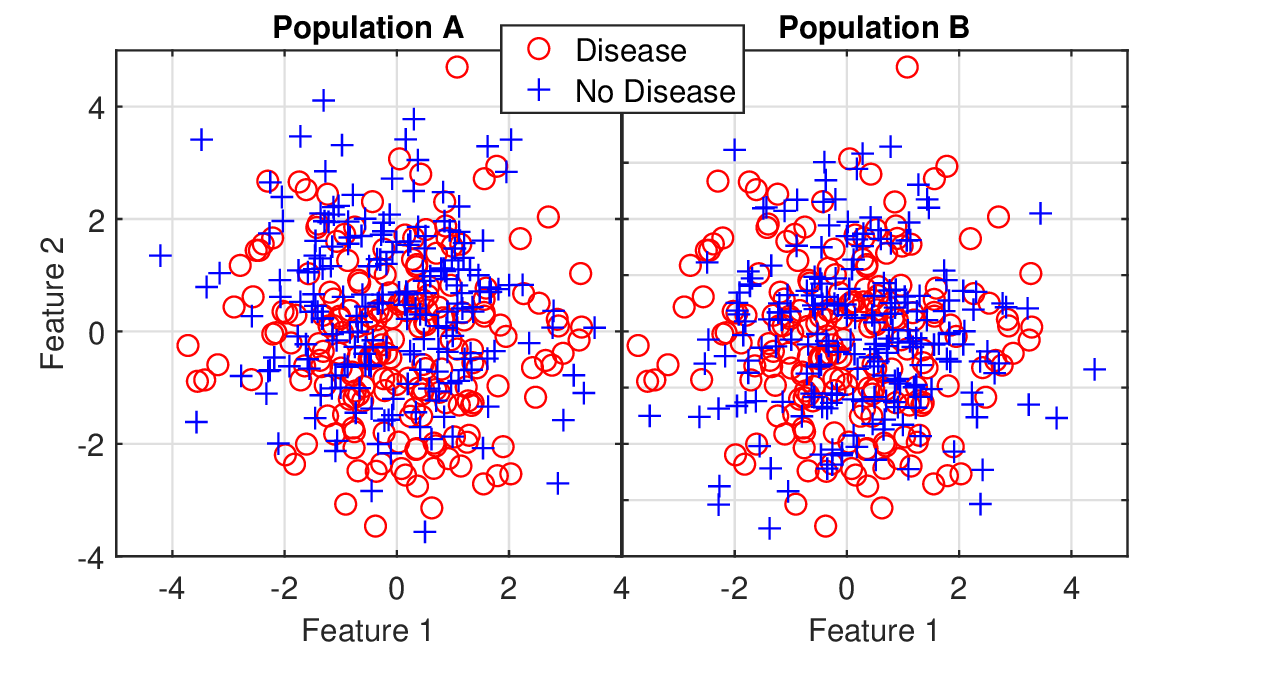}}\\
	\subfloat[][]{\label{subfig:Intuition_Phases}\includegraphics[width = \textwidth,trim=1cm 0.9cm 2.6cm 0.9cm,clip]{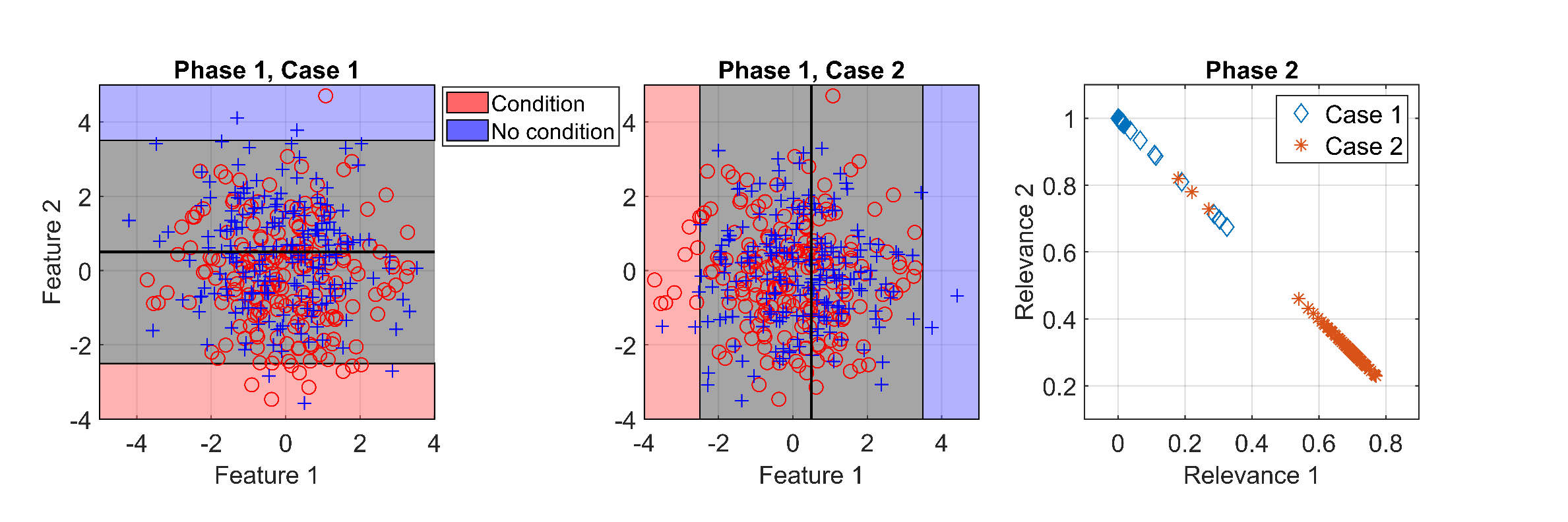}}
	\caption{(a): Two-dimensional representation of the two considered populations (\enquote{A} and \enquote{B}) with two classes (\enquote{Disease} and \enquote{No Disease}). (b): Phase 1 - Case 1 (top) and Case 2 (center), with uncertainty in the estimation of a discriminative hyperplane as a gray band. The bottom panel shows Phase 2 classification of the relevances estimated in Case 1 (diamonds) and Case 2 (asterisks), relative to 100 classifiers trained in each Case.}
	\label{fig:Intuition}
\end{figure*}
Let us consider the study of the impact of a specific disease across two different populations. Consider the example in Figure \ref{subfig:Intuition}. The data consist of two features and two sets of labels. The first label set indicates the presence or absence of a specific disease (\enquote{Disease} or \enquote{No Disease}), and the second label set provides membership to population \enquote{A} and \enquote{B}. The two classes (\enquote{Disease} or \enquote{No Disease}) significantly overlap in both populations. We train a classifier that provides feature relevances, on each population. Therefore, we are effectively treating the classification task as a subspace learning problem. 

Feature 2 is more informative in population \enquote{A} and feature 1 in population \enquote{B}. In order to consider the variability of the results, a number of classifiers is trained on each population.
From each classifier in each population, we obtain a set of relevance vectors. However, due to the overlap between classes in both populations, the accuracy of all classifiers is low and the empirical distribution of the relevance of the features shows high variability. The uncertainty on the discriminative hyperplane in Case 1 and Case 2 is shown as a grey region in Figure \ref{subfig:Intuition_Phases}, left and central panels. In the proposed methodology, we refer to the process of training multiple classifiers on populations \enquote{A} and \enquote{B} as Phase 1 - Case 1 and Phase 1 - Case 2, respectively. The results of Phase 1 - Case 1 and Phase 1 - Case 2 are sets of feature relevance for the same classification task over the two different populations. 

To capture the difference between the relevance sets, a new classifier can be trained on the relevance vectors obtained from Phase 1 - Case 1 and Case 2, using as labels their membership to either population. The input for this classifier is presented in the right panel of Figure \ref{subfig:Intuition_Phases}. We refer to this process as Phase 2. By performing this classification, we are looking for the features that differ the most in the classification task across the two distinct populations.

It is important to note that the input space of Phase 2 is given by the relevances of the original features. By definition, relevances sum to one and thus live in a $(d-1)$-dimensional simplex, where $d$ is the dimension of the feature space. In the over-simplified scenario presented in this example, there are only two original features. This implies that the two relevance vectors lie on the line with vertices $[1 ~0]$ and $[0 ~1]$: the one-dimensional simplex. The separation vector $[-1 ~1]$ in Phase 2 identifies the original features that differ the most across Case 1 and Case 2.

The purpose of this synthetic experiment is to provide insights into the approach designed in this study. Real-world applications can be more complicated, both in terms of the number of features and overlap between classes, as will be shown in Section \ref{sec:Experimental_Settings}. A formal derivation of the methodology is given in Section \ref{sec:Double_Subspace_Learning_Theoretical_Motivations}.

\section{Theoretical Motivations}
\label{sec:Double_Subspace_Learning_Theoretical_Motivations}

Consider a typical two-class classification problem: a dataset $\cD = \{(\x_i, \ell(\x_i)) \:|\: i = 1,\dots,n, ~ \x_i\in \RR^d ,\ell(\x_i) = \{1, 2\}\}$ of points sampled from the two classes $c_1$ and $c_2$, identified by labels $\ell(\bm{x}_i)$. 
Denote by $\cX$ and $\cL$ the sets of training inputs and labels, that is:
 $\cX = \{\x_i \in \RR^d \:| i = 1,\dots,n\}$ and $\cL = \{\ell(\x_i) \in \{1,2\} \:| i = 1,\dots,n\}$.

In the following, we assume that the class-conditional distributions for classes $c_1$ and $c_2$ are multivariate spherical Gaussians, centred at $\bm{\mu}_1$ and $\bm{\mu}_2$  with the same covariance matrix $\mathbf{\Sigma} = \nu^2 \I$. In particular:
\begin{equation}
	p(\x|\bm{\theta}_k) = \cN(\bm{\mu}_k,\bm{\Sigma});~ \bm{\theta}_k = \{\bm{\mu}_k,\bm{\Sigma}\},     
\end{equation}
where $k=1,2$, $\bm{\mu}_1 = \bm{0}$, $\bm{\mu}_2 = t \a$ and $\a$ is a unit directional vector ($\|\a \| = 1$) identifying the separation direction between means $\bm{\mu}_1$ and $\bm{\mu}_2$. In the above $t \ge 0$ is a parameter representing the separation between the classes.
Having the class-conditional distributions, we describe the probabilistic model generating dataset $\cX$ (without considering labels) as a flat  Gaussian mixture with components
$p(\x|\bm{\theta}_1)$ and $p(\x|\bm{\theta}_2)$:
\begin{equation}
p(\x) = \frac{1}{2} \left[p(\x|\bm{\theta}_1) + p(\x|\bm{\theta}_2)\right]
\end{equation}
By considering the whole dataset $\cX$ as \emph{iid} generated, the mean of the mixture model $p(\x)$ is $\bm{\mu}_T = \frac{1}{2}t\a$.

The estimated covariance matrix of dataset $\cX$ under the distribution given by $p(\x)$ is:
\begin{equation}\label{eq:EstimatedSigma}
	\C = \mathrm{Cov[\x]}_{p(\x)} = \nu^2 \I + \frac{t^2}{4}\a\a^\top,
\end{equation}
where we used $\bm{\Sigma} = \nu^2 \I$.

Consider two Cases for the separation vector $\a$. In Case 1, $\a = \e_1 = (1,0,...,0)^\top$, the first vector of the standard basis, while in Case 2, $\a$ is expressed in terms of a rotation by an angle $\alpha < \pi/2$ in the $(\e_1,\e_2)$-plane. This makes the relevant classification subspace (to be discovered by the metric tensor $\bm{\Lambda}$) 2-dimensional and identical to the $(\e_1,\e_2)$-plane. In order to test the 

In our setting, it is reasonable to approximate $\bm{\Lambda}$ by the covariance matrix $\bm{C}$ in eq. \eqref{eq:EstimatedSigma}. Indeed, given the construction of the classification tasks, the eigenvectors of $\bm{\Lambda}$ and $\bm{C}$ should coincide.

We think  of Case 1 and Case 2 as representing classification tasks related to the same phenomenon, but grounded in two different populations.
Applying subspace learning algorithms to both cases, we would expect the respective eigenvectors to be at an angular distance of $\alpha$ in the $(\e_1,\e_2)$-plane.
Collecting the relevance vectors in both Cases, we study their separation in Phase 2, where we identify relevant features in the classification across the two Cases. Experimental evidence suggests that this separation (Phase 2) is larger than the individual separations in both Cases of Phase 1, for small values of the separation parameter $t$.

\subsection{Phase 1 - Case 1}\label{subsec:P1C1}
 When the separation parameter $t$ is sufficiently large, the dominant eigenvector of metric tensor $\bm{\Lambda}$ is aligned with the true separation direction $\a = \e_1$ (in Case 1). The other eigenvectors are perpendicular to $\a$. However, at the same $t$, the variance of dataset $\cD$ will be larger along direction $\a$ than along any other direction. Thus, the dominant eigenvector of the covariance matrix $\bm{\Sigma}_T$ (estimated by $\C$, Eq. \ref{eq:EstimatedSigma}) is also $\a$. Thus, assuming eigenvectors of norm $1$, the eigen-decomposition of metric tensor $\bm{\Lambda}$ and estimated covariance matrix $\C$ provide the same eigenvectors, with the dominant one being $\a = \e_1$. The only difference between $\bm{\Lambda}$ and $\C$ is that the first is positive semi-definite, while the second is positive definite. This means that, while the eigenvectors are in principle the same, the associated eigenvalues are different. In particular, the eigenvalues of $\bm{\Lambda}$ might be zero. Given this similarity, by investigating the eigen-spectrum of $\C$, we are able to provide upper-bounds for the eigenvalues of $\bm{\Lambda}$. Also, since the relation between eigenspectra and relevance of the metric tensor is well-known, we are able to convert the upper bounds on the eigenvalues into the corresponding relevances.

Given that $\a = \e_1$, the eigenvectors of $\C$ form the standard basis. We can estimate the normalized eigenvalues $\tilde \sigma_k$ of $\C$ corresponding to eigenvectors $\bm{v}_k = \e_k$ ($k=1,\dots,d$). 
The eigenvalues are normalised to sum to one, $\sum_{k=1}^d \tilde \sigma_k = 1$. The relevance vector elements (the diagonal terms of the metric tensor) are:
\begin{equation}
	r_j = \sum_{k=1}^d \tilde\sigma_k (\bm{v}_{j,k})^2
\end{equation}
Before plugging in vector $\e_1$, let us first recover the eigenvalue associated with the dominant eigenvector in the metric tensor. Since the separation of the classes occurs only along vector $\a$, we know that the dominant eigenvector of $\bm{\Lambda}$ has to be $\a$. Thus, we need to solve the eigenvalue problem $C\a = \sigma_1 \a$. By eq. \eqref{eq:EstimatedSigma} we have: 
\begin{equation}
	\C\a = \left(\nu^2 \I+ \frac{t^2}{4} \a\a^\top\right)\a = \nu^2\left( 1 + \frac{t^2}{4\nu^2}\right)\a = \sigma_1 \a
\end{equation}
where we used $\a\a^\top = \|\a\|^2_2 = 1$. 
By imposing $\a = \e_1$, we are able to derive the values of all other eigenvalues, by simply solving the eigenvalue problem of the associated eigenvectors perpendicular to $\a=\e_1$, i.e. the standard basis $\e_2,\dots,\e_d$, finding:
\begin{equation}
	\C\e_j = \left(\nu^2 \I + \frac{t^2}{4} \a\right)\e_j = \nu^2 \e_j
\end{equation}
This result is true for all $j \neq 1$, due to the fact that $\e_j\e_k = \delta_{jk}$, where $\delta_{jk}$ is the Kronecker symbol and  $\delta_{jk} = 1$ if and only if $j = k$, otherwise it is $0$. We now need to compute the normalized eigenvalues $\tilde\sigma_j$ by dividing each $\sigma_j$ by the sum over all dimensions:
\begin{equation}
\sum_{k=1}^d \sigma_k = \sigma_1 + \sum_{k=2}^d \sigma_k = \nu^2\left(d + \frac{t^2}{4\nu^2} \right)
\end{equation}

Let us first introduce the notation $\gamma^2(t)$ for the Kullback–Leibler divergence between two multivariate Gaussian distributions with means $\bm{\mu}_1 = \bm{0}$ and $\bm{\mu}_1 = t\a$ and covariance matrix $\bm{\Sigma} = \nu^2 \I$:
\begin{equation}
\gamma^2(t) = \int \cN(\bm{\mu}_1,\Sigma) \log \left[\frac{\cN(\bm{\mu}_1,\Sigma)}{\cN(\bm{\mu}_2,\bm{\Sigma})}\right] \mathrm{d}\x = \frac{t^2}{2 \nu^2}
\end{equation}

Taking advantage of this notation, the normalized eigenvalues take the form:
\begin{equation}
	\tilde \sigma_1  = \frac{1 + \hat{\gamma}(t)}{\xi(t)}; \qquad \tilde \sigma_{j\neq 1} = \frac{1}{\xi(t)},
\end{equation}
where $\hat{\gamma}(t) = \frac{\gamma^2(t)}{2}$ and $\xi(t) = d +\hat{\gamma}(t)$ carries the dependency on the separation factor $t$ via the KL divergence $\gamma(t)$.
From this and the fact that the eigenvectors are the standard basis vectors, we derive the relevance vector $\bm{\rho}^{(1)}$ for Case 1 as:
\begin{align}
	\label{r1}
	\bm{\rho}^{(1)} = \frac{1}{\xi(t)}\left[ 1 + \hat{\gamma}(t), 
	1,  
	\dots, 1\right]^\top.
\end{align}
\subsection{Phase 1 - Case 2}\label{subsec:P1C2}
In Phase 1 - Case 2, we consider $\a$ to be $\R\textbf{e}_1$, where we have introduced a rotation matrix $\R$ as follows: 
\begin{center}
	$\R = \left[
            \begin{array}{c | c}
               \begin{array}{c  c}
                 \cos(\alpha) & -\sin(\alpha) \\
                 \sin(\alpha) & \cos(\alpha)
                \end{array} & \bm{0} \\ [10pt]
              \hline
              \bm{0} & \bm{1} 
            \end{array}
    	\right] $
\end{center}
From this, vectors $\a$ and $\b$ ($\b$ perpendicular to $\a$) are the first two columns of $\R$ (and the first two eigenvectors of $\bm{\Lambda}$), while the $d-2$ remaining eigenvectors are still the vectors of the remaining standard basis:
\begin{align}
\begin{split}
	\a &= \R\e_1  = \left[ \cos(\alpha), \sin(\alpha), 0, \dots, 0 \right]^\top\\
	\b &= \R\e_2  = \left[ -\sin(\alpha), \cos(\alpha), 0, \dots, 0 \right]^\top\\
	   \e_{j>2} &= \R\e_{j>2}
 \end{split}
\end{align}
However, the eigenvalues do not change with respect to Phase 1 - Case 1, given the separation $t$, eigen-decomposition is rotationally invariant. From this we can estimate the relevance of each feature in Case 2 and collect them in the corresponding relevance vector:
\begin{equation}
\bm{\rho}^{(2)} = \frac{1}{\xi(t)}\left[1 + \hat{\gamma}(t) \cos^2(\alpha), 1 + \hat{\gamma}(t) \sin^2(\alpha), \bm{1}\right]^\top,
\end{equation}
where $\bm{1}$ is the $(d-2)$-dimensional vector with only ones.
It is worth mentioning that, following the methodology used in GMLVQ, we enforce $Tr(\bm{\Lambda}) = 1$. This condition is necessary in order to avoid identifiability problems in training.

\subsection{Phase 2}\label{subsec:P2}
In Phase 2, we can compute the separation between the relevance vectors of Case 1 and 2 by realizing that the direction of separation is always along the vector $\bm{\rho}^{(2)} - \bm{\rho}^{(1)}$. Since all relevances are larger than $0$ and lower bounded by the inverse of $d + \gamma^2(t)$, we refer to this separation as \textit{pessimistic}.
It can be shown that the \textit{pessimistic} separation is then:
\begin{equation}\label{Pessimistic_Separation}
\varepsilon_P = \|\bm{\rho}^{(2)} - \bm{\rho}^{(1)}\| = \sqrt{2} \left(\frac{\hat{\gamma}(t)}{\xi(t)}\right) \sin^2(\alpha)
\end{equation}

\subsubsection{Stationary Separation, Optimistic Scenario}\label{subsec:StationarySep}
In order to define the separation, we make use of the  formulation for the stationarity of the relevance matrix of a two-class problem, described in \cite{biehl2015stationarity}. 
The stationary relevance matrix is rank $1$, by construction of the data set, with only one eigenvector equal to the dominant vector for the separation. Furthermore, the relevance vectors are imposed to sum to one.
In Case 1, the separation vector is $\e_1$, thus:
\begin{equation}
	\bm{\Lambda} = \e_1 \e_1^\top
\end{equation}
and the only non-zero relevance is $\lambda^{(1)}_1 = 1$ and so the relevance vector is $\bm{\lambda}^{(1)} = [1, 0, \dots, 0]^\top$.

In Case 2, the only eigenvector of $\bm{\Lambda}$ is $\a = \bm{R} \e_1$ and
\begin{center}
	$\R = \left[
            \begin{array}{c | c}
               \begin{array}{c  c}
                 \cos^2(\alpha) & \cos(\alpha)\sin(\alpha) \\
                 \cos(\alpha)\sin(\alpha)  & \sin^2(\alpha)
                \end{array} & \bm{0} \\ [10pt]
              \hline
              \bm{0} & \bm{0} 
            \end{array}
    	\right] $
\end{center}
so that the relevance vector $\bm{\lambda}^{(2)}$ of Phase 1 - Case 2 is:
\begin{equation}
    \bm{\lambda}^{(2)} = \left[\cos^2(\alpha), ~\sin^2(\alpha),~0,\dots,0 \right]^\top
\end{equation}
The relevances again sum to one. By the same argument adopted in the previous section for the computation of the separation in Phase 2, we can now identify the \textit{optimistic} separation in which all relevances are null except the truly meaningful ones for the classification, finally obtaining the \textit{optimistic} separation:
\begin{equation}
	\label{Optimistic_Separation}
 \varepsilon_O = \|\bm{\lambda}^{(2)} - \bm{\lambda}^{(1)} \| = \sqrt{2}\sin^2(\alpha)
\end{equation}
The dependence on $t$ is dropped, since the separation direction in both Case 1 and 2 is always assumed to be identified correctly.

\subsubsection{Optimistic vs. Pessimistic Separations}\label{subsec:OptvPess}

Consider now the two derived separations for the \textit{pessimistic} and \textit{optimistic} scenarios, in equations \eqref{Pessimistic_Separation} and \eqref{Optimistic_Separation} respectively. Equation \eqref{Pessimistic_Separation} can be rewritten as:
\begin{equation}
	\varepsilon_P(t) =  \sqrt{2} \left(\frac{1}{1 + \beta^2(t)}\right) \sin^2(\alpha)
\end{equation}
where we have introduced the notation:
\begin{equation}
	\beta(t) = \frac{\sqrt{2d}}{\gamma(t)} = \frac{2 \nu\sqrt{d}}{t} = \frac{2 \sqrt{d}}{\tilde{t}}.
\end{equation}
Here $\tilde{t} = t / \nu$ is the separation between classes in units of standard deviations of their conditional distribution.
The proportion between \textit{pessimistic} and \textit{optimistic} separations is then:
\begin{equation}
	\frac{\varepsilon_P(t)}{\varepsilon_O} = \frac{1}{1 + \beta^2(\tilde{t})}
\end{equation}
This equation clearly displays the relationship between optimistic and pessimistic scenarios with the separation value $\tilde{t}$. When $\tilde{t}$ gets larger, $\beta$ tends to $0$ and the $\varepsilon_P(t)$ approaches $\varepsilon_O$ with order of $O(\frac{1}{\tilde{t}^2})$. It also shows that the higher the dimensionality of the input space $d$, the more $\varepsilon_P(t)$ underestimates $\varepsilon_O$.

\subsubsection{Normalized Experimental Separation}
\label{subsec:Exp_on_Synt}
Let us now consider the experimental setup for Phase 2. We assume that we have performed GMLVQ for both Case 1 and 2 for $n=100$ times resulting in $n$ different metric tensors and relevance vectors per Case. In order to study the separation between the relevance vectors in Phase 2, we consider the $n$ sets of relevance vectors in each Case to be independent. The sets containing the relevance vectors in Case 1 and 2 are:
\begin{equation*}
	\mathcal{R}^{(1)} = \{ \r^{(1)}_i| i=1,\dots,n\}, \quad \mathcal{R}^{(2)} = \{ \r^{(2)}_i| i=1,\dots,n\}
\end{equation*}
and their estimated means:
\begin{equation*}
	\langle\r^{(1)}\rangle = \frac{1}{n} \sum_{i=1}^n \r^{(1)}_i, \quad \langle\r^{(2)}\rangle = \frac{1}{n} \sum_{i=1}^n \r^{(2)}_i
\end{equation*}
The estimated mean separation vector can then be approximated by $\overline\r = \langle\r^{(2)}\rangle -  \langle\r^{(1)}\rangle$. Its norm provides the \textit{experimental} separation: $\varepsilon_E = \| \overline\r \| = \| \langle\r^{(2)}\rangle -  \langle\r^{(1)}\rangle \|$. 
Normalizing the separation vector provides the unit norm separation vector $\hat \r = \overline\r / \varepsilon_E$. This allows for the computation of the projected relevance vectors onto the separation direction and the estimate of their projected variability:
\begin{equation}
	p^{(k)}_i = \left( \r^{(k)}_i - \langle\r^{(k)}\rangle \right)^\top \hat \r, \quad \varsigma^{(k)} = \sqrt{\frac{1}{n-1} \sum_{i=1}^n \left[ p^{(k)}_i\right]^2},
\end{equation}
where $k=1,2$.
By taking the average of the two terms $\varsigma^{(1)}$ and $\varsigma^{(2)}$, we derive a single measure for the variability of the separation $\varepsilon_E$. Furthermore, we can define a new measure that quantifies the separation in units of variability:
\begin{equation}
\label{eq:ExperimentalSep}
	\delta_E = \frac{\varepsilon_E}{\overline \varsigma}, \quad \overline \varsigma = \frac{\varsigma^{(1)} +\varsigma^{(2)}}{2}
\end{equation}

\section{Experimental Settings}
\label{sec:Experimental_Settings}
In the following sections, we apply the proposed methodology to two data sets. The first one is a synthetic example where the separation vector between classes in Phase 1 - Case 1 is perpendicular to the one in Phase 1 - Case 2. In the following, these vectors are identified by $\a_1$ and $\a_2$, respectively. 
By smoothly varying the angle $\alpha$ between the two vectors and the separation parameter $t$ between classes in each Case, we can estimate how the separations presented in \ref{subsec:P2} and \ref{subsec:StationarySep} vary with respect to these parameters.
Finally, the methodology is applied to a real-world dataset.
We first show the results with GMLVQ as the base learner and then with SVM\footnote{The GMLVQ MATLAB toolbox can be found at \url{ https://www.cs.rug.nl/~biehl/gmlvq.html}. All results in this work have been obtained with the latest version of the toolbox (v3.1) and default parameters. The \texttt{fitclinear} function of MATLAB R2024b with lasso regularization has been used in all results concerning SVM. Again, default parameters were adopted.}.
\begin{figure*}[ht!]
	\centering
	\subfloat[][]{\label{subfig:AUC_withT_GMLVQ}\includegraphics[width = \textwidth,trim = 0.5cm 0.8cm 0.7cm 0.9cm,clip]{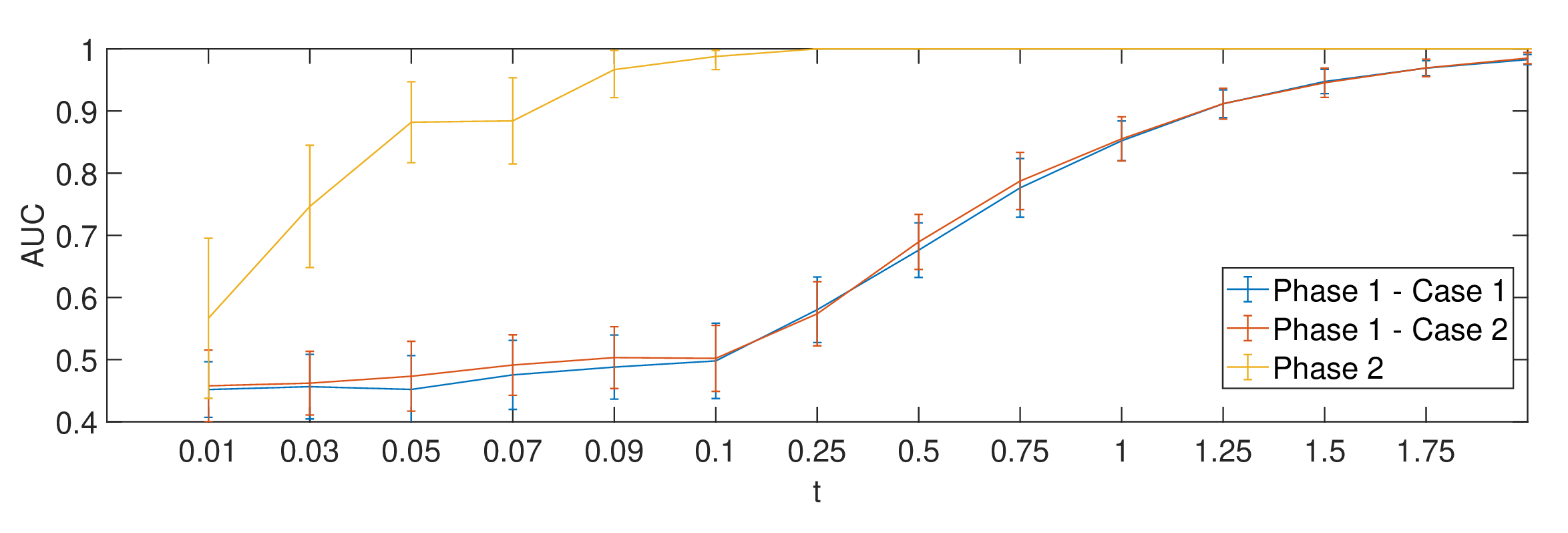}}\\
	\subfloat[][]{\label{subfig:AUC_withT_SVM}\includegraphics[width = \textwidth,trim = 0.5cm 0.6cm 0.7cm 0.9cm,clip]{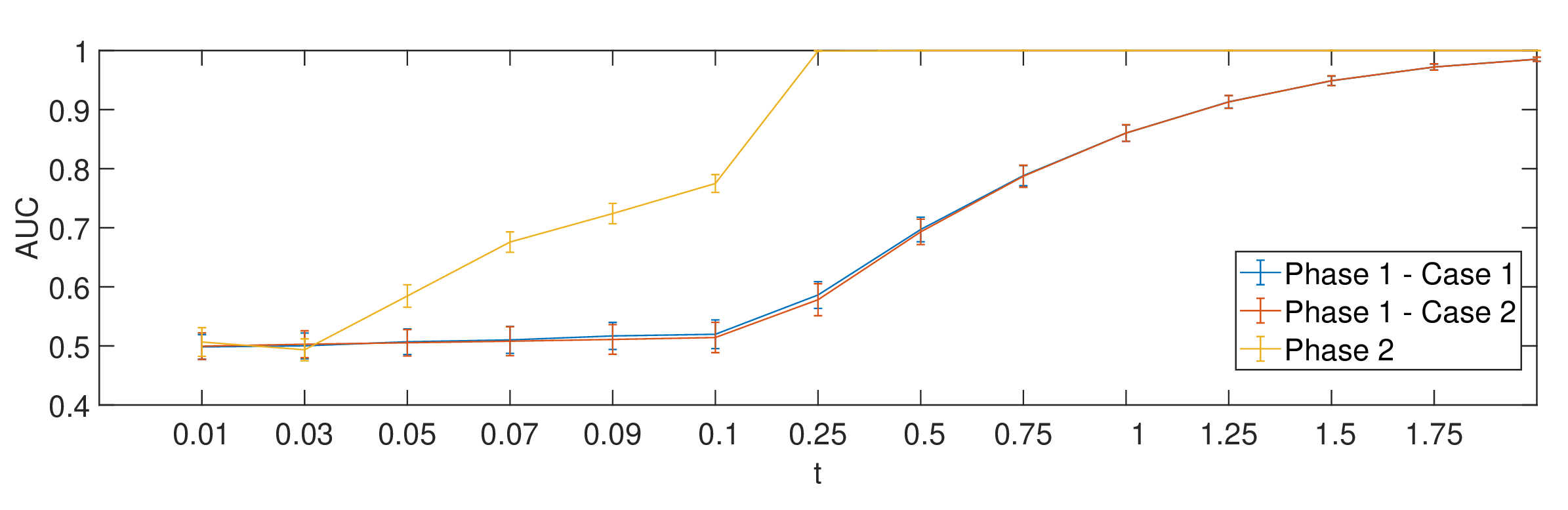}}
	\caption{AUC of classification tasks for Phase 1 - Case 1 (cyan), Case 2 (orange) and Phase 2 (yellow) at varying separation $t$ with GMLVQ (panel \ref{subfig:AUC_withT_GMLVQ}) and SVM \ref{subfig:AUC_withT_SVM}. The vertical bars show the mean and standard deviation of the results over 100 trials in each Phase and Case.}
	\label{fig:AUC_withT}
\end{figure*}

\subsection{Synthetic data set}

In order to have an estimate of the dependency of Discriminative Subspace Emersion (DSE) from the separation of the two classes in each Case, we first consider the two separation vectors to be orthogonal. In this view, we define the separation vector for Phase 1 - Case 1 as $\a_1$ and the one for Phase 1 - Case 2 as $\a_2$:
\begin{align*}
\a_1 &= [1, 1, 0, 0, \frac{1}{2}, \frac{1}{2}, 0, 0, 0, 0, 0, 0, 0, 0, 0, 0, 0]\\
\a_2 &= [0, 0, 1, 1, 0, 0, \frac{1}{2}, \frac{1}{2}, 0, 0, 0, 0, 0, 0, 0, 0, 0]
\end{align*}
As described in Section \ref{sec:Double_Subspace_Learning_Theoretical_Motivations}, the mean of class $1$ in both Case 1 and 2 is on the coordinate origin, while the mean of class $2$ lies on vectors $\a_1$ and $\a_2$ for Case 1 and 2, respectively. The variance of each class is $\nu^2 = 1$ and the covariance matrix $\bm{\Sigma} = \nu^2~\mathbf{I}$. To construct the datasets, $n=500$ \emph{iid} samples are drawn from the Gaussian distributions $\cN(\bm{\mu}_1 = \bm{0},\Sigma)$ and $\cN(\bm{\mu}_2 = t \a_1,\Sigma)$ for Phase 1 - Case 1 and. The same setting is considered for Phase 1 - Case 2 but with class means $\bm{\mu}_1 = \bm{0}$ and $\bm{\mu}_2 = t \a_2$.
The orthogonality between $\a_1$ and $\a_2$ can be easily verified by evaluating $\a_1^\top \a_2 = 0$. However, the two vectors are constructed so that the sets of features relevant in Case 1 and Case 2 are disjoint. 
We believe that the identification of feature relevance is achievable even when the overlap between classes in both Case 1 and Case 2 hinders the individual classification tasks. We test this by having the separation parameter $t$ within the values $0,01$ to $2$ and studying how the effectiveness of classification in Phase 2 differs from Phase 1 - Case 1 and Case 2.

In Phase 1, for both Case 1 and 2, the two classes are generated by sampling independently from the corresponding Gaussian distributions. This operation is performed 100 times in order to obtain measures of variability of the relevant quality metrics. Each time $i$, in both Cases, the base learner (GMLVQ or SVM) is trained over the randomly generated samples. The relevance vectors $\bm{r}_i^{(1)}$ and $\bm{r}_i^{(2)}$ are recovered for Case 1 and 2 and collected in datasets $\cR^{(1)}$ and $\cR^{(2)}$.
We define two new sets of labels $\cL^{(1)} = \{1,\dots,1 \}$ and $\cL^{(2)} = \{2,\dots,2 \}$ that assign feature relevance vectors to either Case 1 or Case 2. By calling $\cD^{(1)} = \cR^{(1)} \times \cL^{(1)}$ and $\cD^{(2)} = \cR^{(2)} \times \cL^{(2)}$ the sets of relevance vectors with their respective labels for both Case 1 and 2, the set $\cD = \cD^{(1)} \cup \cD^{(2)}$ forms the training set for Phase 2. Having obtained 100 different realizations of relevance vectors for both Case 1 and Case 2, we apply the chosen base learner in order to identify the relevant features that separate the two classes of relevance vectors. Training the classifier on this new set yields the final set of feature relevance vectors for Phase 2.
\subsubsection{\textbf{GMLVQ as base learner}}\label{subsubsec:Synthetic_GMLVQ}

\begin{figure*}[h!]
	\centering
	\subfloat[][]{\label{fig:S_FIM} \includegraphics[width=\textwidth]{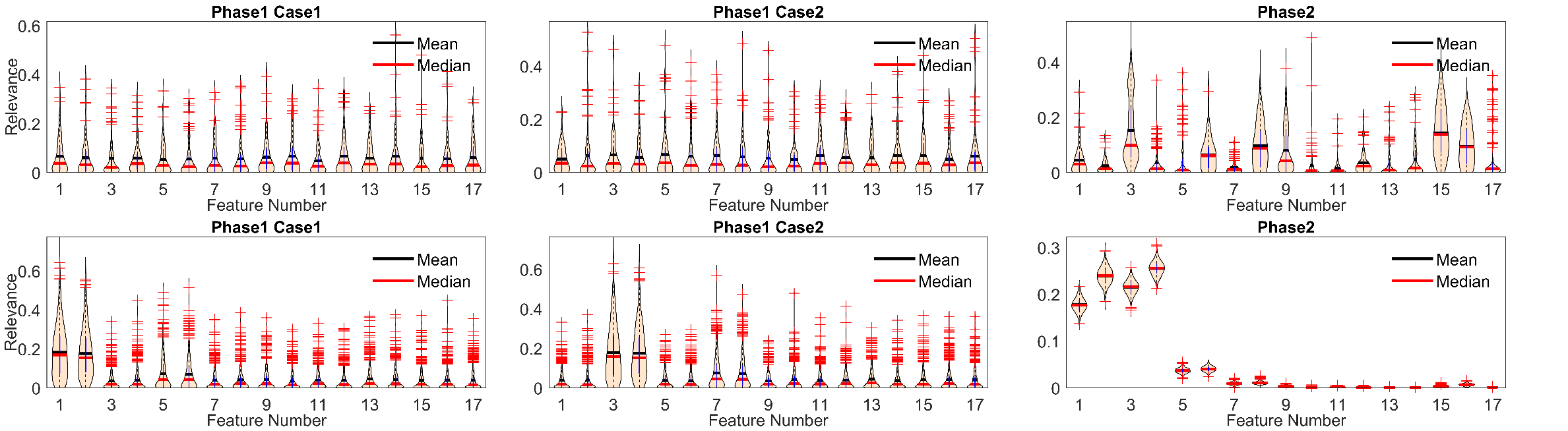}}\\
	\subfloat[][]{\label{fig:S2_FIM} \includegraphics[width=\textwidth,trim={3.2cm 0cm 3cm 0.4cm},clip]{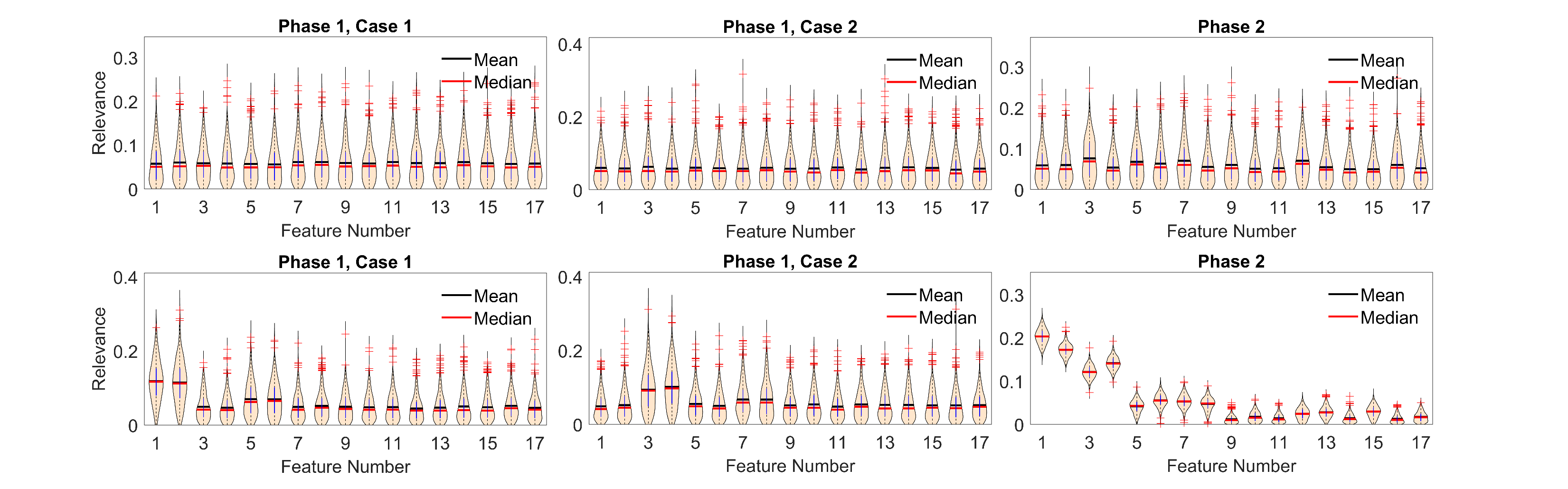}}\\
	\subfloat[][]{\label{fig:S_Plt_p1}\includegraphics[width=\textwidth]{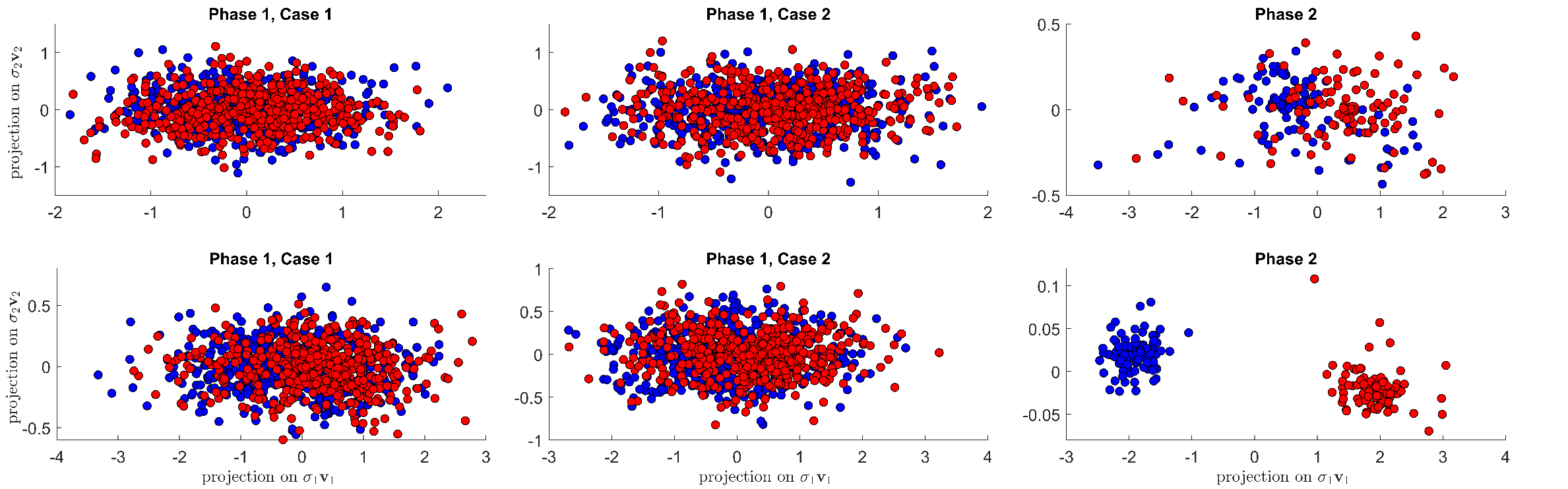}}\\
	\caption{Feature relevance vectors with GMLVQ (a) and SVM (b) as base learners, two-dimensional embedding of samples obtained with GMLVQ (b) for different values of $t$ for the synthetic data for DSE. In each plot, top row $t = 0.01$ and bottom row $t = 0.25$; Column 1: Phase 1 - Case 1; Column2: Phase 1 - Case 2; Column 3: Phase 2.}\label{fig:AllGMLVQ}
\end{figure*}
Figure \ref{fig:AUC_withT} shows the AUC of Phase 1 - Case 1 and 2 and Phase 2 (cyan, 
orange and yellow respectively) for different values of the separation parameter $t$. Both the means and standard deviations over the 100 realizations of Phase 1 and 2 of GMLVQ are shown here. The same is presented for SVM in panel (b). Phase 2 always outperforms Phase 1 in terms of AUC, for any separation value $t$, indicating that when comparing the same classification task over two different populations, subspace learning in the space of relevances recovers cumulative faint signals from the original tasks.

Figure \ref{fig:S_FIM} presents the feature relevance vectors for Phase 1 - Case 1 (left column), Case 2 (central column) and Phase 2 (right column) for two different values of class separation: $t=0.01$ (top row) and $t=0.25$ (bottom row). We show violin plots of distribution of relevances for each feature over the 100 runs of GMLVQ in each Phase/Case. Red crosses report outliers while red and black bars identify the median and mean respectively, of the estimated distributions, for each feature. The same information is reported in \ref{fig:S2_FIM} for SVM as base learner.

Training the classifiers in Phase 2 results in the relevance vectors shown in the right column of the same figures. As expected, when $t$ increases (overlap decreases), the performance of both Cases in Phase 1 increases, and the correct features responsible for the separations are identified as relevant (features $\{1,2,5,6\}$ for Case 1 and $\{3,4,7,8\}$ for Case 2). In Phase 2, some relevance is assigned at further features, for sufficiently high overlap. However, for slightly larger separation values ($t=0.25$), a much clearer scenario is identified in Phase 2, despite the persistence of some noise in Phase 1. Roughly all $8$ features designed to differ the most across the two populations (in reference to the same classification task) are indeed recovered.

This phenomenon is also visible when projecting the classes samples on the two-dimensional embeddings given by the first two dominant learnt eigenvectors obtaine dwith GMLVQ (scaled by the square root of the corresponding eigenvalues) for each step of DSE, as can be seen for an instance of the process in Figure \ref{fig:S_Plt_p1}. Again, as in Figure \ref{fig:S_FIM}, top row is for $t=0.01$ and the bottom row for $t=0.25$. While the top row shows still some overlap in Phase 2 between the relevances estimated in Case 1 and Case 2, it is absent in Phase 2 for $t=0.25$ (bottom row).

While the high performance in Phase 2 does not necessarily imply a high fidelity in separating classes in Phase 1, it still gives an indication about the confidence in determining differently important features in the two cases. 
We stress that the classification tasks are different in Phase 1 and 2. Phase 1 works in the original feature space and performs classification over the Condition of interest in two disjoint populations, while Phase 2 operates on the space of relevances and the task is to distinguish between the two populations. We choose to report average ROC and AUC as the classical way of estimating performance for a given classification task.

\subsubsection{\textbf{Comparison of Phase 2 separations}}
\begin{figure*}[th!]
	\centering
    \includegraphics[width=\textwidth]{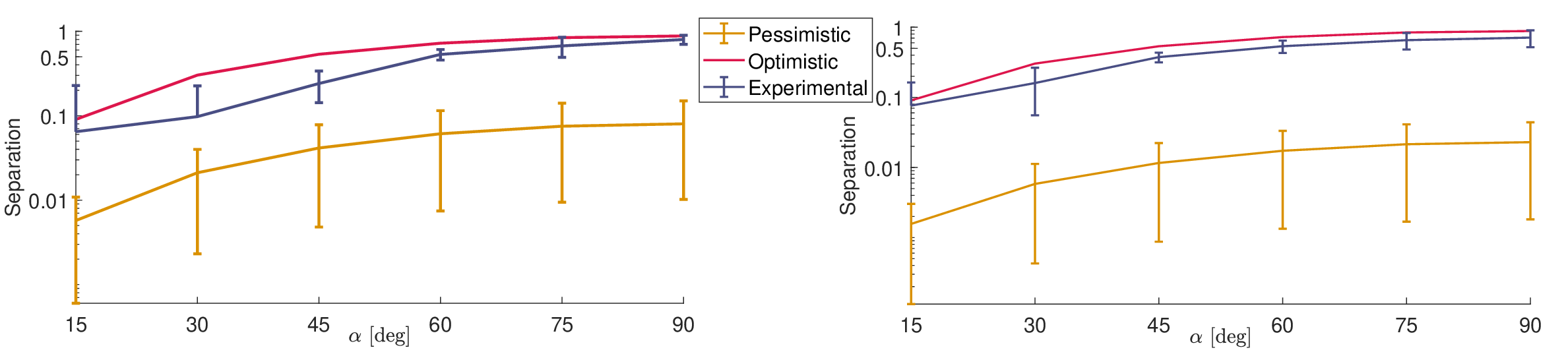}
	\caption{Pessimistic (yellow) , Optimistic (magenta) and Experimental (purple) separations (in log scale) for two different dimensions $d$; $d = 5$ (left panel) and $d = 20$ (right panel).}
	\label{Analytical_Sep}
\end{figure*}
In the following, we evaluate and test the different separation measures described in Section \ref{sec:Double_Subspace_Learning_Theoretical_Motivations} on the synthetic dataset. The proposed separation measures are the \textit{optimistic} separation (Eq. \eqref{Optimistic_Separation}), the \textit{pessimistic} separation (Eq. \eqref{Pessimistic_Separation}), and the \textit{experimental} separation (Eq. \eqref{eq:ExperimentalSep}). 

To test for variability with the angle between separation vectors $\a_1$ and $\a_2$ we define angle $\alpha$ within the interval $[0^{\circ}, ~90^{\circ}]$. Since a certain degree of smoothness is expected, we consider angles within the angular space at a regular distance of $15^{\circ}$. Also, being the \textit{pessimistic} separation dependent on the data dimensionality $d$, we test the relationship between these metrics at two distinct dimensionalities: $d = 5,~10,~15,~20.$
Results for $d = 5,~20.$ extremes are shown in Figure \ref{Analytical_Sep} left and right panel, respectively, at varying angle $\alpha$.
In each panel, yellow bar represents the pessimistic, magenta the optimistic and purple the experimental separations. The error bars on pessimistic and experimental separations represent the standard deviation from the mean when computed at different $t$ values. For every dimensionality and angle, the experimental separation is always upper bounded by the optimistic separation and lower bounded by the pessimistic one.
As expected from Eq. $\eqref{Pessimistic_Separation}$, the pessimistic separation decreases when the dimensionality increases.

\subsection{Adrenal tumours data set}

Benign adrenal tumours are incidentally discovered in $3-7\%$ of adults and are associated with cortisol excess in up to $50\%$ of cases \cite{bancos2021approach}. Cortisol excess is associated with an increased risk of cardiometabolic disease \cite{fassnacht2023european}, and the objective of the study is to use the 24-hour urine steroid metabolome to assess the cardiometabolic risk profile of patients with benign adrenal tumours. This dataset has $1240$ prospectively recruited patients with benign adrenal tumours who underwent measurement of the cortisol and related steroid hormone metabolites in 24-hour urine samples (24-hour urine steroid metabolome analysis \cite{prete2022cardiometabolic}). Each observation has 17 steroid features, which are denoted as $S_1$ through $S_{17}$ in the experiments.
\begin{figure*}[ht!]
	\centering
    \includegraphics[width=\textwidth]{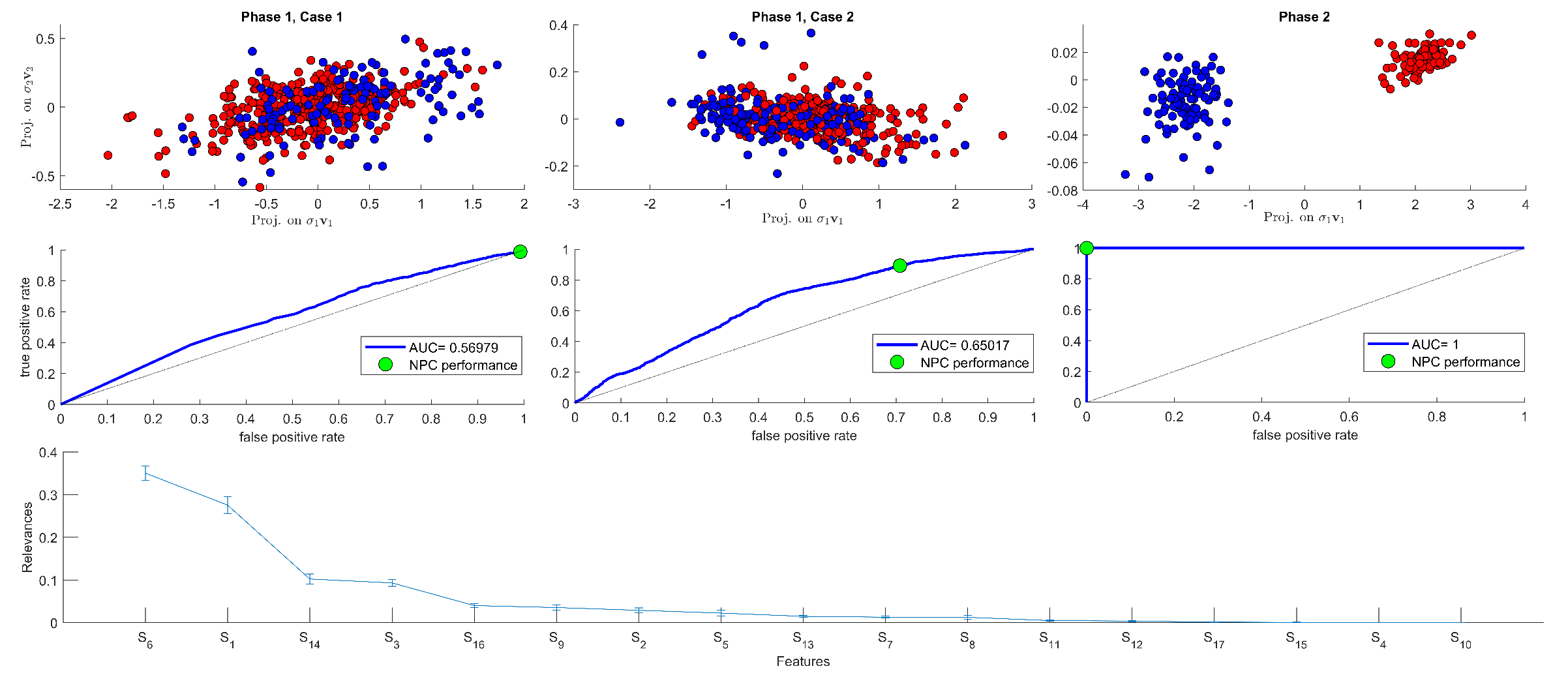}
	\caption{Results of the experiment for adrenal tumours data on two considered populations, population A and population B, for the given health condition. (Top row): Two-dimensional embeddings in Phase 1 - Case 1 / Case 2 and Phase 2 (left to right); (Middle row): ROC of Phase 1 - Case 1 / Case 2 and Phase 2 (left to right); (bottom panel): Sorted feature relevance vectors in Phase 2.}
	\label{Diabetes}
\end{figure*}
Patients are distributed across two populations
(``A'' and ``B'') characterized by different degrees of cortisol excess, and may or may not have one of $3$ different health conditions: Conditions 1, 2 and 3. To show the potential of the methodology we present here the results for the classification of patients with or without Condition 1, across populations A and B. 

The distribution of data across the condition and the two populations is imbalanced. We use random undersampling to provide balanced sets of training data and avoid imbalanced data issues. We thus create 100 sets of balanced classes for both Population A and B. In our terminology, classification over the health condition for population A is Phase 1-Case 1, over population B is Phase 1-Case 2. On each balanced set, for each Case in Phase 1, we train GMLVQ and obtain the corresponding relevance vectors for the classification task. The two sets of relevance vectors are then used in Phase 2 to identify relevant features for the classification over the condition across the two populations.

Figure \ref{Diabetes} shows the results of the experiments conducted on the 24-hour urine steroid metabolome of patients with benign adrenal tumours to predict clinical characteristics. While Phase 1 shows considerable overlap between clinical characteristics, Phase 2 accurately identifies steroid features, that according to the current understanding of steroid production pathway \cite{greaves2014guide}, are driving the difference between two groups with different degrees of cortisol excess, but with the same condition.
It should be noted that the goal of DSE is not to build a more efficient classifier for a given task, but to recover a subspace where the task differs the most if performed over different populations.

\section{Discussion and Conclusion}
\label{Conclusion}
In this paper, we presented Discriminative Subspace Emersion (DSE) as a novel methodology for identification of relevant features in the classification task over two considered populations. It operates in two Phases, first learning important features for the classification task in Population A and B (Cases 1 and 2), then finding the subspace that best distinguishes between the task over the two populations (Phase 2). 
The workings of the methodology have been shown on a synthetic data set carefully designed to evaluate the variability of the results with respect to the inherent degrees of freedom of the methodology. These are the severity of overlap between classes in Phase 1 and the angle between separation directions in Phase 1 - Case 1 and 2. Extensive experiments across synthetic data sets indicate that DSE can identify the feature relevance effectively even in situations of high overlap between classes of Phase 1. 

The experiments were carried out with Generalized Matrix Learning Vector Quantization (GMLVQ) as base learner. However, other methodologies could have been used as well, as long as the classification results in Phase 1 provide additional information about relevant features for the classification (relevances). In this sense, methods such as Random Forests \cite{breiman2001random,598994}, logistic regression or Support Vector Machine (SVM) could be used as base learners. In additional analysis we performed the same experiments with SVM as base learner. Over the synthetic data set, the application of either learner provided similar results in terms of identified different relevant features for the classification across the two considered populations, proving that DSE can be implemented with other subspace learning algorithms as base learners. 

The methodology is applied to a biomedical case study, where the distinction between patients with a certain condition is required over two degrees of cortisol excess. The recovered steroid features are meaningful tracers of differences across the two populations for the given clinical condition.
We believe DSE to be a powerful tool, and to best of our knowledge the first one, in the detection of faint signals across multiple binary classifications over different populations.

\bibliography{Bibliography}

\end{document}